\documentclass[lettersize,journal]{IEEEtran}
\usepackage{amsmath,amsfonts}
\usepackage{array}
\usepackage{textcomp}
\usepackage{stfloats}
\usepackage{url}
\usepackage{verbatim}
\usepackage{graphicx}
\usepackage{cite}
\usepackage{bm}
\usepackage{pifont}
\usepackage{float}
\usepackage{hyperref}
\usepackage{algorithm}
\usepackage{algpseudocode}
\usepackage{booktabs}
\usepackage{multirow}
\usepackage{amssymb} 

\IEEEoverridecommandlockouts                              

\begin{document}


\title{\LARGE \bf
MRNaB: Mixed Reality-based Robot Navigation Interface using Optical-see-through MR-beacons
}

\author{Eduardo Iglesius$^{\dag 1}$, Masato Kobayashi$^{\dag 2*}$, Yuki Uranishi$^{2}$, Haruo Takemura$^{2}$
\thanks{$^{1}$ Graduate School of Information Science and Technology, The University of Osaka, 
$^{2}$ D3 Center, The University of Osaka, 
$^{\dag}$ Equal Contribution, 
$^*$ corresponding author: kobayashi.masato.cmc@osaka-u.ac.jp
}
}

\maketitle
\thispagestyle{empty}
\pagestyle{empty}

\begin{abstract}
Recent advancements in robotics have led to the development of numerous interfaces to enhance the intuitiveness of robot navigation. However, the reliance on traditional 2D displays imposes limitations on the simultaneous visualization of information. Mixed Reality (MR) technology addresses this issue by enhancing the dimensionality of information visualization, allowing users to perceive multiple pieces of information concurrently. This paper proposes the Mixed Reality-based Robot Navigation Interface using an Optical-see-through MR-beacons (MRNaB), a novel approach that uses MR-beacons created with an ``air tap'', situated in the real world. This beacon is persistent, enabling multi-destination visualization and functioning as a signal transmitter for robot navigation, eliminating the need for repeated navigation inputs. Our system is mainly constructed into four primary functions: ``Add'', ``Move'', ``Delete'', and ``Select''. These allow for the addition of MR-beacons, location movement, its deletion, and the selection of MR-beacons for navigation purposes, respectively. To validate the effectiveness, we conducted comprehensive experiments comparing MRNaB with traditional 2D navigation systems. The results show significant improvements in user performance, both objectively and subjectively, confirming that the MRNaB enhances navigation efficiency and user experience. For additional material, please check: \url{https://mertcookimg.github.io/mrnab}
\end{abstract}


\section{Introduction}
The Robot Operating System (ROS) \cite{journal:ros} has been a pivotal framework in the evolution of robot navigation interfaces. It has enabled researchers to devise interfaces that are not only easy to use but also rich in functionality \cite{ROSUnder, ROSWheelchair, ROSNeuro, ROShip}. The development of these robot navigation interfaces has been crucial in advancing the field of robotics, making complex operations more accessible to users.
However, the reliance on 2D displays in current robot navigation interfaces presents certain limitations. Users are often required to interpret a map generated through Simultaneous Localization and Mapping (SLAM) while managing the robot's real-world movements. This dual focus can reduce the efficiency of operations, as it demands frequent shifts of gaze and attention between the map and the real environment \cite{journal:arviz}.

\begin{figure}[t]
    \centering
    \includegraphics[width=1 \columnwidth]{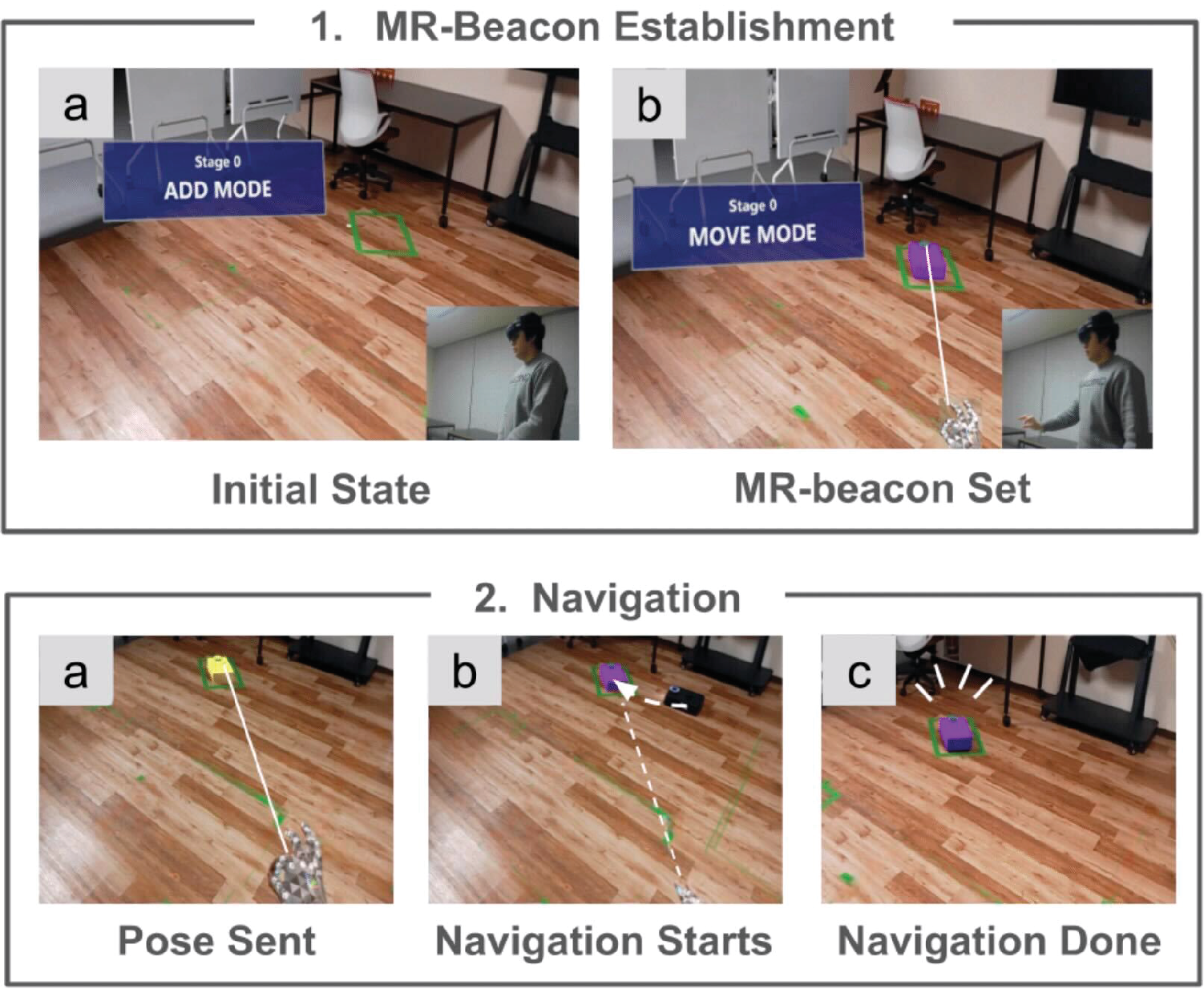}
    \caption{MRNaB Concept. (1-a) shows the initial state where there is no MR-beacon on the floor. Next, (1-b) MR-beacon will be set to a certain location on the floor where even after leaving the project or restarting it, the MR-beacon will still be there and be used for robot navigation. (2-a) For navigation, the beacon just needs to be clicked by the user, then (2-b) the navigation will start and (2-c) the robot will move to the desired place.}
    \label{fig:SysConcept}
\end{figure}

Recently, many researchers have focused on MR \cite{Salvato2022mr, Devo2022mr, AR4Robot} as a tool for enhancing human-computer interaction by offering a more intuitive and immersive interface. MR significantly improves the user experience by providing a unified view of virtual and real-world elements, thereby minimizing the need for attention shifts and enhancing operational effectiveness\cite{Fan2023mr, Penco2024mr, Zhang2023mr}.
Numerous studies have introduced innovative robot navigation interfaces through the use of an MR device called Hololens~2 \cite{journal:arviz, eyeoperation, journal:drag, journal:gaze}. However, specifying a navigation destination within a specific location, especially during delivery processes in households or companies, can be challenging. Existing interfaces encounter limitations in two main areas: Firstly, existing interfaces use markers like arrows to set the robot's destination, but users often struggle to visualize the robot's status at the destination, especially in space-constrained areas. This difficulty can cause the user to fail to navigate the robot to its intended location. Secondly, users need to repeatedly specify the robot's pose for frequently visited destinations, which diminishes efficiency. 

This paper introduces MRNaB, a novel approach for robotic navigation by employing a persistent MR-beacon, which can be created by ``air tap'', that mirrors the actual shape of the robot and retains its position even after the project is restarted, useful for visualization in multiple repetitive destinations. Fig.~\ref{fig:SysConcept} shows the concept of the system. This approach not only simplifies the task of visualizing the robot's destination in navigation but also streamlines the process of setting destinations that are regularly used. In order to implement the system, we provide the user with 4 main functions which are ``Add'' to add the MR-beacon from the floor, ``Move'' to move the location of the MR-beacon, ``Select'' to select the MR-beacon for robot navigation, and ``Delete'' to delete the MR-beacon that has been made.
We also implemented a database system, which eliminates the need for users to repeatedly specify the robot's location. This feature is particularly beneficial for environments that require repetitive navigation to the same place, such as households, especially for delivery tasks. Our contributions are summarized as follows:
\begin{itemize}
\item This paper proposes MRNaB, a mixed reality navigation interface with an MR-beacon shaped like the robot, which can be used for multiple destinations by utilizing the ``air tap'' gesture, improving destination visualization and enhancing navigation efficiency.
\item MRNaB supported persistent goal poses, allowing users to set frequently used destinations, beneficial for repetitive navigation tasks in various settings.
\item We conducted experiments comparing MRNaB with traditional 2D display interfaces, demonstrating our advantages in user experience and operational efficiency in robot navigation.
\end{itemize}

This paper consists of six sections including this one.
Section II explains the related works.
Sections III and IV introduce proposed system design and system implementation.
Section V shows the experimental results to confirm the usefulness of the proposed method.
Section VI concludes this paper.

\section{Related Works}

\subsection{2D Robot Navigation Interface}

In the field of robotic navigation, 2D interfaces have been instrumental in simplifying the complexities of robot control and monitoring. Simulation tools \cite{gazebo, MRDS, webots, Web-based, web-based3, web-based4} have been pivotal in this development. However, these 2D robotic simulation interfaces face a limitation: commands in the simulation do not directly influence real-world robots. 

RViz \cite{rviz}, web-based live operation interface \cite{web-based2}, and Kachaka application address this gap by not only visualizing robot information but also impacting real-world robots by utilizing the 2D map or 3D map created by SLAM shown in the screen of the computer. Nonetheless, even this interface encounters specific challenges:

$\bullet$ {\it Seamless Attention}: Users interacting with the program must shift their attention between the screen and the real robot, leading to a disconnect from the physical robot/environment and making it less effective for certain tasks.

$\bullet$ {\it Real World Spatial Understanding}: A 2D map, generated by SLAM using the robot's Lidar, provides only a rough estimation of the environment based on the Lidar's height. This limitation becomes apparent in typical household structures, which do not conform to uniform shapes like cubes, thus complicating spatial comprehension from a 2D perspective. 3D maps, though can give the understanding of the space, but is not as clear as the real-world itself and the creation of 3D maps is resource-intensive, especially for dynamic environments with frequent changes.

Our research aims to bridge the identified challenges by augmenting user visual capabilities through the integration of real-world information and 2D maps generated by SLAM for robotic navigation. This approach addresses several key issues: it enables immediate impact on real-world robots, mitigates resource constraints, enhances spatial understanding, and eliminates the need for users to constantly shift their focus between the display and the real-world environment. This approach also adapts effectively to dynamic environments by utilizing a 2D map for robot movement and real-world information for user operations. Since the 2D map represents navigable space, small environmental changes do not require remapping, ensuring the robot can continue operating without disruption. Additionally, by enhancing spatial understanding through HoloLens 2, our system improves the user’s ability to interpret and adapt to the environment.

\subsection{MR Robot Navigation Interface}
\begin{table}[t]
\centering
\caption{Comparison to Other Interfaces in Robotic Navigation}
\scalebox{1}{ 
\begin{tabular}{lcccc}
\toprule
\multirow{2.5}{*}{\textbf{Index}} & \multicolumn{2}{c}{\textbf{2D}} & \multicolumn{2}{c}{\textbf{MR}} \\
\cmidrule(lr){2-3}  \cmidrule(l){4-5} 
               & I1      & I2    & I3 & MRNaB \\ 
\cmidrule(lr){1-1} \cmidrule(lr){2-2} \cmidrule(lr){3-3} \cmidrule(lr){4-4} \cmidrule(l){5-5} 
Affecting Real World Robot             &                 \ding{55}  &            \checkmark           & \checkmark & \checkmark\\ 
Seamless Attention             &          \checkmark         &                \ding{55}        & \checkmark & \checkmark\\ 
Real World Spatial Understanding              &     \ding{55}            &               \ding{55}       & \checkmark & \checkmark\\ 
Multi-Destination Visualization            &         \checkmark         &          \checkmark            & \ding{55} & \checkmark\\ 
Destination Persistency              &     \checkmark             &               \checkmark       & \ding{55} & \checkmark\\ 
\bottomrule
\end{tabular}
} 
\small

Note: I1 - 2D Interface by Simulation \cite{gazebo} - \cite{web-based4}. \quad I2 - 2D Interface by Live Operation \cite{rviz} - \cite{web-based2}. \quad I3 - Other MR Interface \cite{journal:arviz, journal:drag, journal:gaze}, \cite{surrogate} - \cite{map2}. 
\label{tab:robotic_navigation_interfaces}
\raggedright
\end{table}
In the field of MR for robotic navigation, research initiatives have explored innovative methods to enhance human-robot interaction. Studies include using Hololens~2 to visualize robot information through occlusions \cite{thruWall} and employing markers and signs to indicate robot motion intent, safety distances, and interaction guidelines \cite{intent, ihandai}. Specialized interfaces for wheelchair robots have been developed to display intent \cite{wheelchair, wheelchair2}, and multi-user interfaces facilitate simultaneous interactions with multiple robots \cite{multiuser}. 
Lee et al. also explore the eye gaze field by using it to operate the robot manually \cite{eyeoperation}. Kot et al. combine a joystick and Hololens~2 to operate the robot manually from other places \cite{joystickoperation}. However, these studies are for manual navigation.

Walker et al.\mbox{\cite{surrogate}} proposed similar idea by using MR surrogates for teleoperating robots with a joystick, aiding users in visualizing the robot's destination. However, their method requires additional hardware, focuses on teleoperation and only shows the last waypoint for multiple waypoints, potentially confusing users about each waypoint's significance and its relation to the overall route, particularly in household settings.

This paper focuses on the MR interface for autonomous navigation.
For autonomous navigation, one notable example is ARviz \cite{journal:arviz}, which enhances 2D maps created by SLAM through the use of Hololens~2, enabling visualization of information provided by Rviz. Robot navigation is facilitated by arrows created by ``air tap'' hand gestures to show the goal of the navigation. Some also use waypoints for navigation goals \cite{Drone, holo-spok, sar}. Chen et al. made an approach involving an interface that employs holograms and ``drag and drop'' interface for visualization and navigation by 3D map created by SLAM \cite{journal:drag}. Wu et al. use the 2D map with waypoints interface to navigate the robot \cite{map2}. Zhang et al. also explore the utilization of eye gaze and head movement for robot navigation and robot arm control \cite{journal:gaze}.

Despite the advancements these interfaces offer, they share common challenges:

$\bullet$ {\it Multi-Destination Visualization}: Users are unable to visualize the actual condition or status of the robot's destination for multiple destinations. This limitation makes navigation difficult, especially in space-constrained areas, such as household. Displaying multiple markers in a confined space can lead to confusion, as it may not be immediately clear how each marker relates to the robot’s intended movement.

$\bullet$ {\it Destination Persistency}: Once the application is restarted or the destination is set, previously set destinations are lost, necessitating the recreation of new destination points each time the application is run.

Our research proposes to overcome these challenges by enhancing the navigation interface by representing the robot's goal pose with an object that matches the robot’s actual dimensions, thus making it easier for user to visualize where the robot wants to go even for multiple destinations. Our research also proposes implementing a database system that allows users to retrieve previously set destinations upon re-running the program. 

To sum up, the comparison between our interface and other interfaces can be expressed in Table \ref{tab:robotic_navigation_interfaces}.
2D interface despite having multi-destination status visualization and destination persistency, has a problem with seamless attention and real-world spatial understanding since it only uses the map. MR interface addresses this problem. However not being able to visualize the robot's destination for multiple destinations and a feature to navigate to repetitive places make navigation difficult. This research tries to bridge all these challenges by providing all the features provided by both 2D interfaces and other MR interfaces.

\begin{figure*}[t]
    \centering
    \scalebox{1}{
    \includegraphics[width=\textwidth]{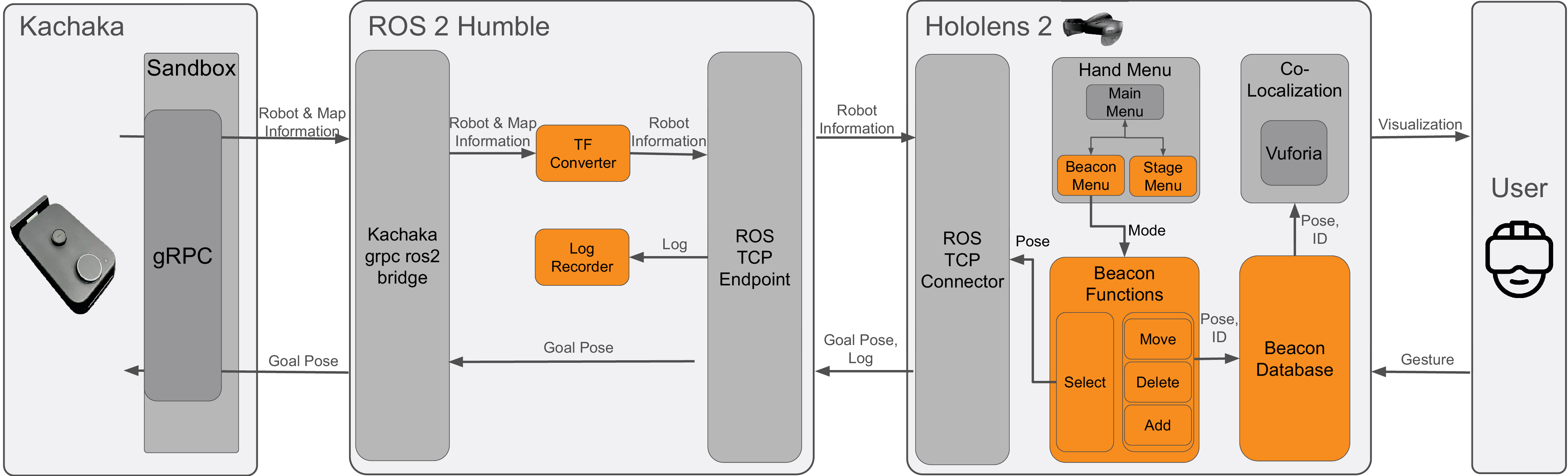}}
    \caption{System Design Diagram. The user interacts with Hololens~2 by using gestures to access the hand menu and interact with the MR beacon to access the beacon functions. Our system also has a database that saves the information (Pose and ID) of the robot and uses it during the first co-localization. Hololens 2 receives robot data while sending the log and goal pose to the ROS 2 side. ROS 2 has a log recorder and TF converter to convert robot and map information to only robot information. Goal pose will be sent directly using the Kachaka gprc ROS~2 bridge to the ``Kachaka'' robot.}
    \label{fig:SysDesign}
\end{figure*}
\begin{figure}[t]
    \centering
    \includegraphics[width=\columnwidth]{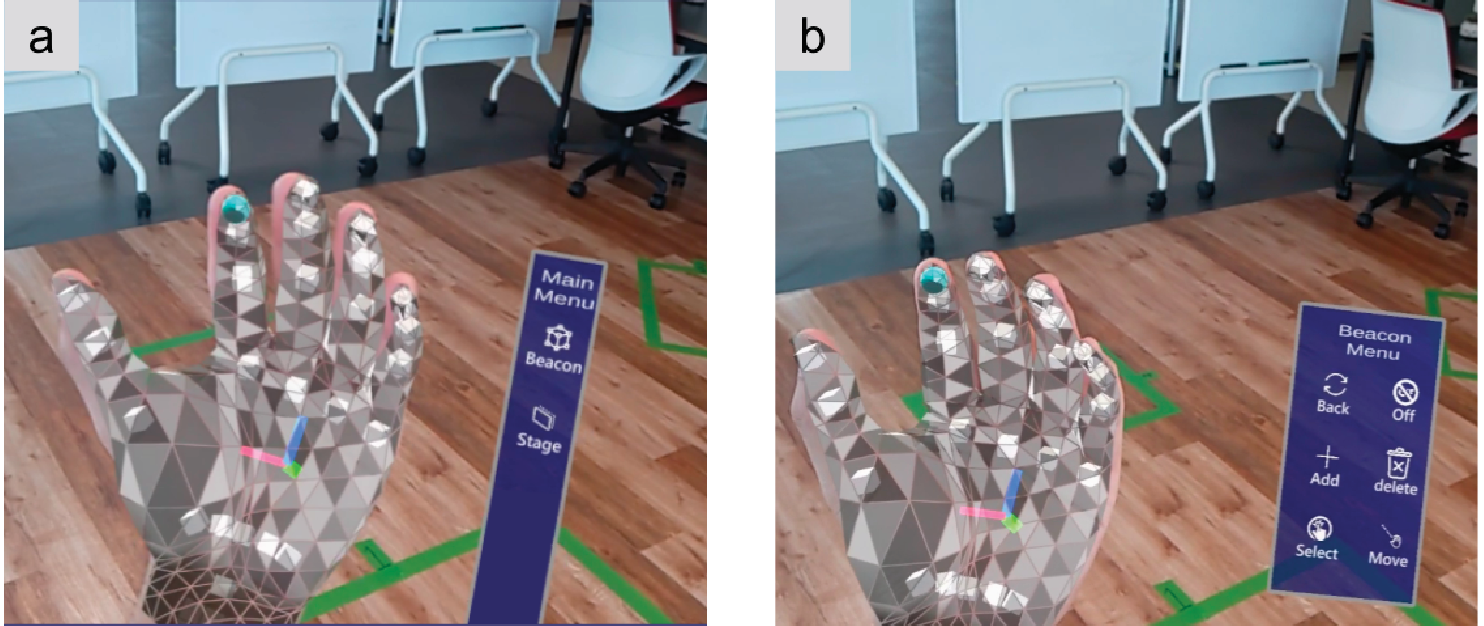}
    \caption{Hand Menu. (a) shows main menu which is shown when user does ”Hand Constraint Palm Up” movement. (b) shows the beacon menu after pressing the beacon button in main menu.}
    \label{fig:HandMenu}
\end{figure}

\section{System Design}

Our system design is illustrated in Fig. \ref{fig:SysDesign}. We employ ROS~2 for interfacing the robot (Kachaka) with the user, while Hololens2 is utilized for user interaction.
In order to connect ROS~2 and Hololens~2, we use ROS-TCP-Connector. Orange-colored components are our novelty component to support our system. Co-localization is done by using vuforia by putting a QR image target on the floor level which resembles the map topic on the robot.
All of the MR-beacons' coordinates will be measured relative to this image target thus making the transformation to ROS-type data easier.

In order to access the MR-beacon to navigate the robot, the hand menu is used by doing the ”Hand Constraint Palm Up” movement. Hand menu is shown in Fig.~\ref{fig:HandMenu}. In the hand menu, there will be beacon button and stage button. Stage button is only used for experiment. Beacon button will then further be divided into 6 buttons which are the back button, off button, add button, move button, select button, and delete button. Back button is used to return to the main menu and off button is used to remove all the functionality of the MR-beacon. The main functions for this system are add, move, select, and delete functions which correspond to their respective buttons.
\begin{figure}[t]
    \centering
    \includegraphics[width=\columnwidth]{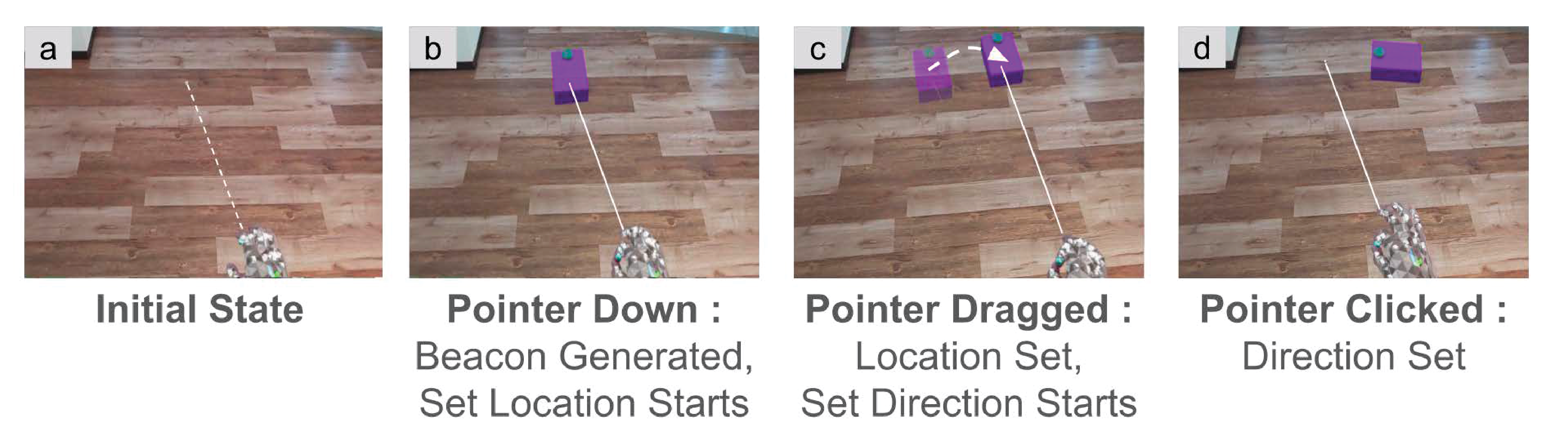}
    \caption{Add Mode Process. (a) shows the initial state where there is no MR-beacon on the floor. (b) When user does the ``air tap'' movement to the floor, MR-beacon will be generated and location setting starts. (c) As user drags the pointer, MR-beacon will follow the pointer. Once ``air tap'' is released, MR-beacon will fix the location and start the direction setting. (d) To fix the direction, user has to do one more ``air tap''.}
    \label{fig:Add}
\end{figure}
\begin{figure}[t]
    \centering
    \includegraphics[width=\columnwidth]{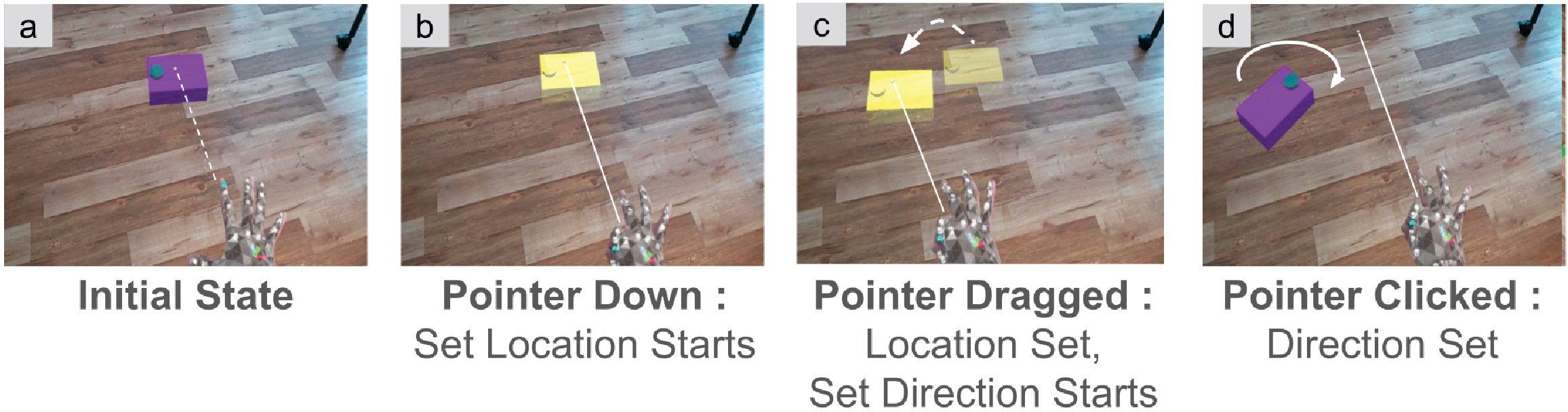}
    \caption{Move Mode Process. (a) shows the initial state where there is an MR-beacon on the floor. (b) When user does the ``air tap'' movement to the MR-beacon, MR-beacon will be generated and the location setting starts. (c) As user drags the pointer, MR-beacon will follow the pointer. Once ``air tap'' is released, MR-beacon will fix the location and start the direction setting. (d) To fix the direction, user has to do one more ``air tap''.}
    \label{fig:Move}
\end{figure}

\subsection{Add Function}

Add function is used to generate MR-beacon from the floor. After pressing add button, user will enter ``Add Mode''. Fig.~\ref{fig:Add} shows the ``Add Mode'' process for MR-beacon. 

Initially, there is no MR-beacon on the floor. When user uses ``air tap'' gesture (pointer down) to the floor, beacon will be generated from the floor and follow the location of the pointer of the user by dragging the pointer. In order to fixate the location of the MR-beacon, user needs to release the ``air tap'' and after that direction setting will start where MR-beacon direction will follow the location of the pointer of the user. To fixate the direction of the MR-beacon, user needs to do ``air tap'' gesture once more (pointer click).

\subsection{Move Function}

Move function is used to move or adjust the pose of the MR-beacon to other poses. After pressing move button, user will enter ``Move Mode'', and user can move the MR-beacon that was generated before. Fig.~\ref{fig:Move} shows ``Move Mode'' process which works similarly to ``Add Mode''. However, instead of doing the ``air tap'' movement to the floor, the user has to do ``air tap'' to the MR-beacon. 

The initial state starts with MR-beacon already on the floor. When user does ``air tap'' gesture to the MR-beacon, the MR-beacon will turn yellow and user can drag the location of the MR-beacon freely on the floor.  In order to fixate the location of the MR-beacon, user needs to release the ``air tap'' and after that direction setting will start where MR-beacon direction will follow the location of the pointer of the user. In order to fixate the direction of the MR-beacon, user needs to do ``air tap'' gesture once more (pointer click).

\begin{figure}[t]
    \centering
    \includegraphics[width=\columnwidth]{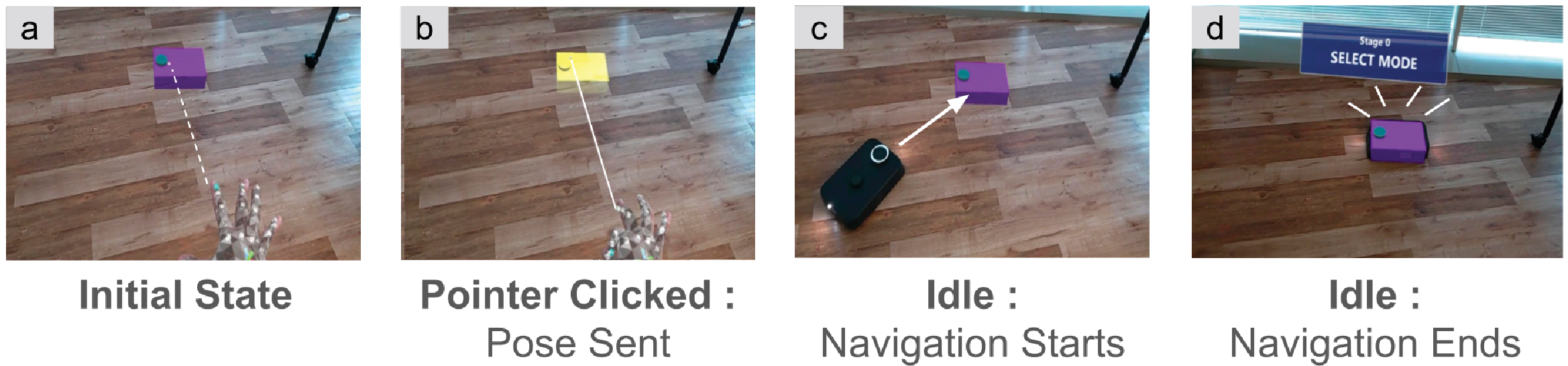}
    \caption{Select Mode Process. (a) shows the initial state where there is an MR-beacon on the floor. (b) By clicking the MR-beacon, MR-beacon's location will be sent to the real robot. (c) Navigation will be started once robot receives the information. (d) Robot reaches the location of the MR-beacon.}
    \label{fig:Select}
\end{figure}
\begin{figure}[t]
    \centering
    \includegraphics[width=\columnwidth]{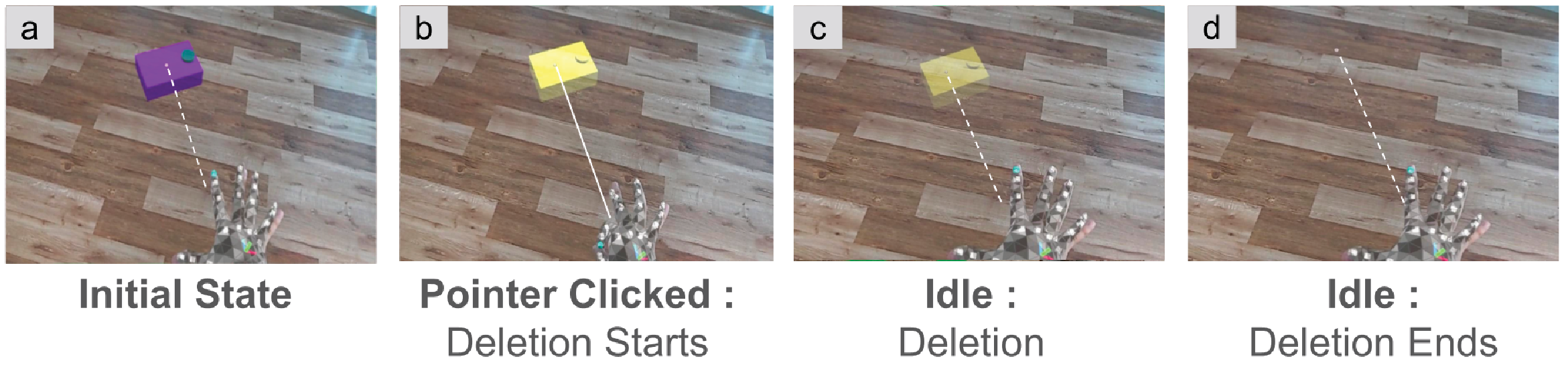}
    \caption{Delete Mode Process. (a) shows the initial state where there is an MR-beacon on the floor. (b) By clicking the MR-beacon on the floor, the deletion process will start. (c) shows the deletion, and (d) shows the state where the beacon is deleted from the floor.}
    \label{fig:Delete}
\end{figure}
\subsection{Select Function}

Select function is used to send the pose of the MR-beacon to the real robot. After pressing select button, user will enter ``Select Mode'', and user can navigate the robot to any of the MR-beacon generated. ``Select Mode'' process is shown in Fig.~\ref{fig:Select}. Initial state starts with MR-beacon on the floor already. By doing ``air tap'' gesture (pointer click) to the MR-beacon, the pose of the MR-beacon will be transformed to be relative to the QR code and be sent to the robot. Robot will then do navigation to the targeted pose.

\subsection{Delete Function}

Delete function is used to delete MR-beacon. After pressing delete button, user will enter ``Delete Mode'' where user can delete any generated MR-beacon on the floor. ``Delete Mode'' process is shown in Fig.~\ref{fig:Delete}. Initial state starts with MR-beacon on the floor. By doing ``air tap'' gesture (pointer click) to any MR-beacon, it will be deleted from the floor. The information of that MR-beacon will also be deleted from the database. 

\section{System Implementation}
\subsection{Communication with ROS~2}
The ROS-TCP-Connector is used for sending and receiving data between Hololens~2 and ROS~2.
On ROS~2 side, ROS-TCP-Endpoint will be made by importing the node file to the workspace and then running it as a node. On Hololens~2 side, ROS-TCP-Connector package will be imported and by specifying the IP address of the ROS-TCP-Endpoint, the connection between Hololens~2 and Unity will be made. The data that will be sent from Unity including the goal pose for robot navigation. The data that will be received by Unity including robot information for co-localization.

\begin{figure}[t]
    \centering
    \begin{minipage}{0.4\textwidth}
        \centering
        \includegraphics[width=0.4\textwidth]{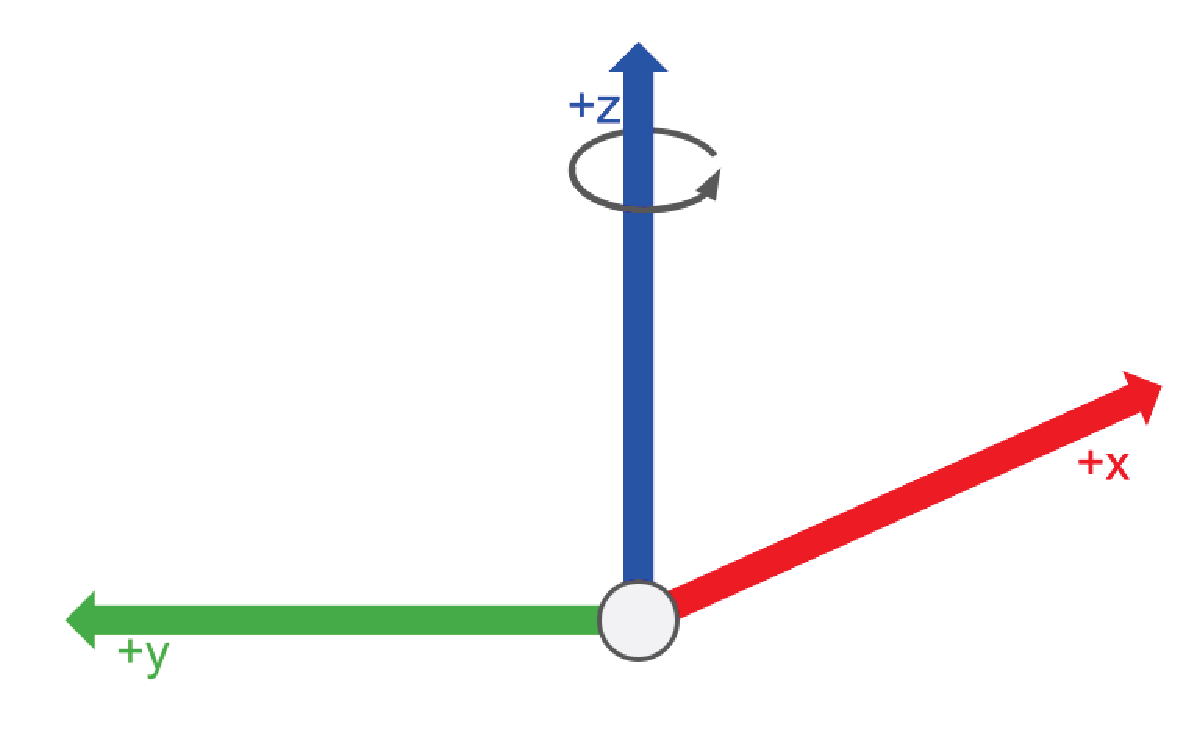}
        \caption{ROS~2 Coordinate System}
        \label{Fig:ros2Coordinate}
    \end{minipage}
    \begin{minipage}{0.4\textwidth}
        \centering
        \includegraphics[width=0.45\textwidth]{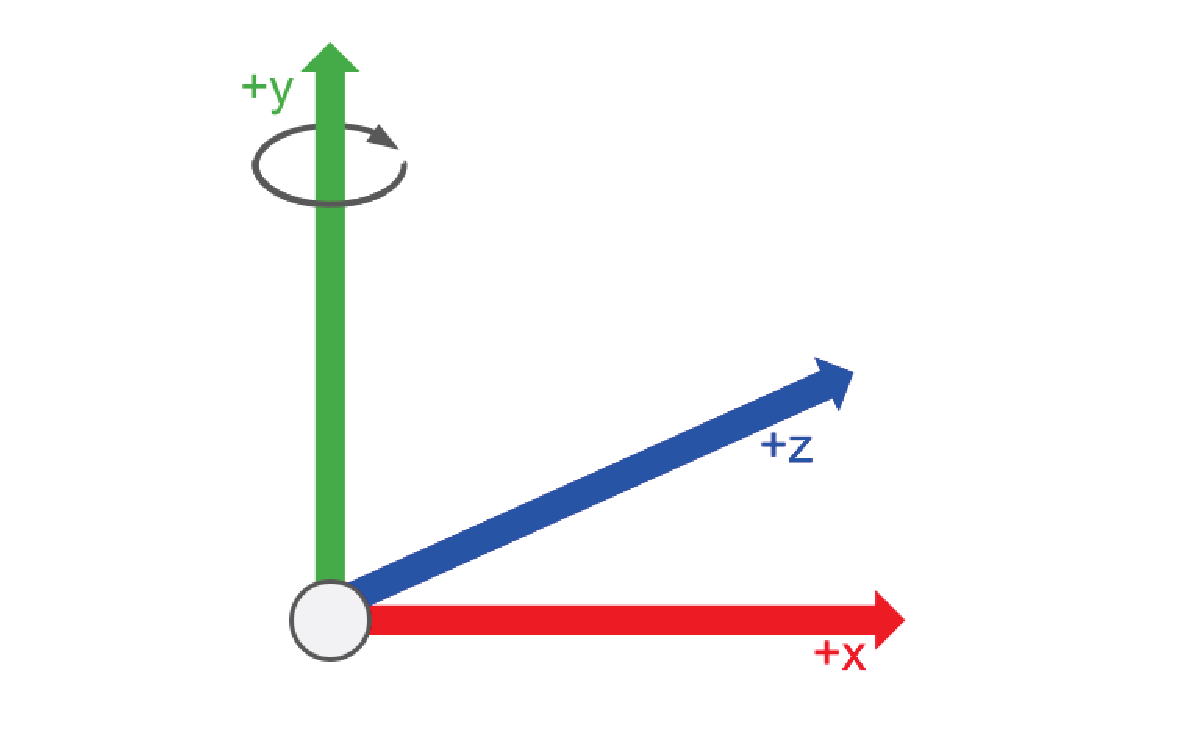}
        \caption{Hololens~2 Coordinate System}
        \label{Fig:UnityCoordinate}
    \end{minipage}
\end{figure}

\subsection{Co-localization}
ROS~2 coordinate system (Fig.~\ref{Fig:ros2Coordinate}) follows the right-handed coordinate system where the positive $x$-axis points forward, positive $y$-axis points to the left, and positive $z$-axis points upward.
Meanwhile, Hololens~2 coordinate system (Fig.~\ref{Fig:UnityCoordinate}) follows the left-handed coordinate system where the positive $x$-axis points to the right, the positive $y$-axis points upward, and the positive $z$-axis points forward.

In order to align the coordinate of ROS~2  and Hololens~2 , co-localization is required. In this system, we use the same co-localization process used by ARviz \cite{journal:arviz} by using vuforia. By using vuforia, real object's prototype will first be put to the vuforia database and then the object will be accessed as a unity object. All of the object made in the unity system will have the parent set to this QR code. We put this QR code to be on the same location of the map location in ROS system, in order to make the transformation from Unity coordinate system to ROS coordinate system easier.

\subsection{MR-beacon}

\subsubsection{MR-beacon Design}
\begin{figure}[t]
    \centering
    \scalebox{0.9}{
        \includegraphics[width=0.9 \columnwidth]{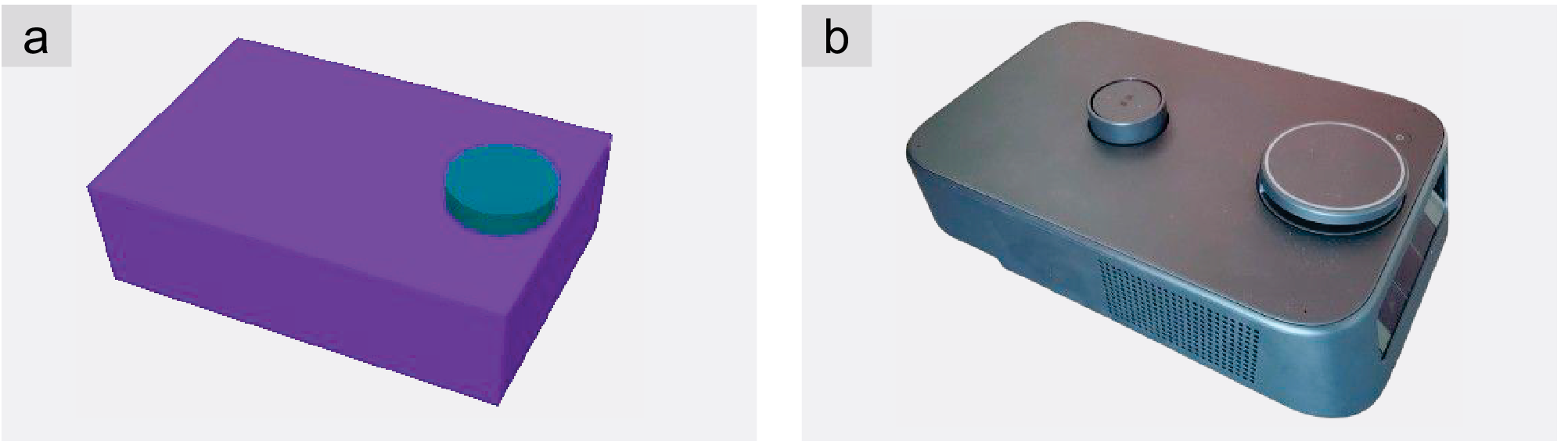}
    }
    \caption{MR-beacon Design. (a) shows the figure of the MR-beacon. (b) shows figure of the real robot (Kachaka).}
    \label{fig:MR-beaconDesign}
\end{figure}

The dimensions of the MR-beacon are designed to mirror those of the real robot we used, as illustrated in Fig.~\ref{fig:MR-beaconDesign}. While the actual shape of the robot is not precisely a block, we simplified its form for the design, disregarding specific elements like the wheels or the camera.

\subsubsection{Add Function}
For MR-beacon add function, since there is no MR-beacon in the first place, we put a transparent plane object at the top of the floor whose origin is located exactly at the QR Code location.  To this object, we then put object manipulator provided by MRTK, while restricting the movement to only $x$-axis and $z$-axis, as well as restricting the rotation in all directions. Once user starts doing the ``air tap'' movement, MR-beacon will be generated on that place. The initial rotation of the MR-beacon will follow the rotation of the Euler angle of the pointer during the first MR-beacon generation.

Once the MR-beacon is generated, the location setting will start. As user drags the transparent floor, the relative location of this MR-beacon will always be the same as the transparent object, thus making it look like the MR-beacon follows the direction of the pointer from the user. Once the location setting is done, this transparent floor will then return to the original place and the object manipulator component of this object will be deactivated. Next, the MR-beacon will undergo direction setting phase. During the direction setting, the rotation of the MR-beacon will follow the location of the pointer on the floor by changing the direction of the object to face toward the location of the pointer.

\subsubsection{Move Function}
MR-beacon move function works similarly to the add function, both are divided into two steps, which are location setting and direction setting. During the location setting, instead of the invisible floor, we put object manipulator provided by MRTK to the MR-beacon itself to relocate the position of the MR-beacon while restricting the movement to only $x$-axis and $z$-axis, as well as restricting the rotation in all direction. Once the location setting is done, we disactivate this object manipulator. The direction setting is also similar as the rotation of the MR-beacon which will follow the location of the pointer on the floor by changing the direction of the object to face toward the location of the pointer.

\subsubsection{Delete Function}
MR-beacon delete function is used to delete the MR-beacon. Once MR-beacon is clicked by the pointer, MR-beacon will then be destroyed from the project.

\subsubsection{Select Function}

For MR-beacon Selection Function, once the MR-beacon is selected, the MR-beacon will then send its current local pose relative to the QR Code to ROS side. Since ROS~2 Humble and Hololens~2 have different coordinate system, before we send the information we transform the information to suit the coordinate system in ROS~2 Humble.

\begin{figure}[t]
    \centering
        \includegraphics[width=\columnwidth]{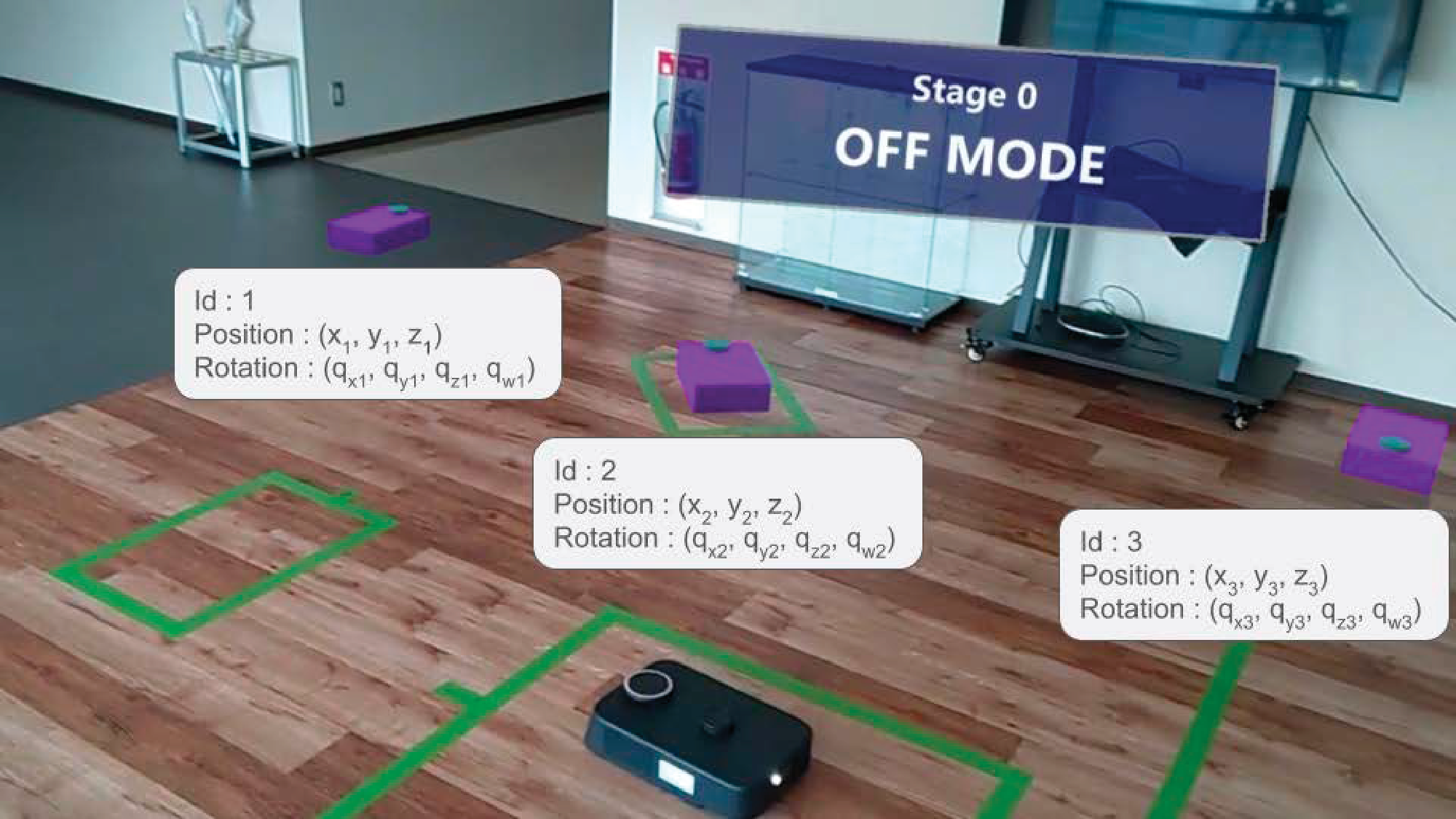}
    \caption{Database Usage in Multi-Destination Visualization and Persistency}
    \label{fig:Multi-Destination}
\end{figure}

\subsection{Database}
\begin{algorithm}[t]
\caption{Pseudocode for Database Management}
\small
\begin{algorithmic}[1]
\While{system is running}
        \State $recentBeaconMode \gets GetBeaconMode()$
        \If{recentBeaconMode is Add}
            \State $database \gets LoadDatabase()$
            \State Add($database$, $ID$, $Pose$) 
        \ElsIf{recentBeaconMode is Move}
            \State $database \gets LoadDatabase()$
            \State Change($database$, $ID$, $Pose$)
        \ElsIf{recentBeaconMode is Delete}
            \State $database \gets LoadDatabase()$
            \State Delete($database$, $ID$)
        \EndIf
        \State SaveDatabase()
\EndWhile
\end{algorithmic}
\end{algorithm}

The database is used to store the information about the MR-beacon. All the information of the MR-beacon that will be stored includes: First, the name of the robot which also can be used as the ID, which is created by using Globally Unique Identifier (GUID). Second, position of the MR-beacon represented by $x$, $y$, $z$ for $x$-axis, $y$-axis, and $z$-axis of the location of the robot relative to the QR code respectively. Third, rotation of the MR-beacon represented by $q_x$, $q_y$, $q_z$, and $q_w$ quaternion of the robot relative to the QR code respectively. 

After the direction setting in the add function, the database will store the information of the established MR-beacon as new data. After the direction setting in the move function, the database will check the moved MR-beacon ID to update the information of the pose. After the delete function is used, the database will remove the item from the database. Lastly, when the user starts the project, if the QR code has been found, all of the stored beacons will be instantiated. The database's pseudocode is shown in Algorithm~1. Fig.~\ref{fig:Multi-Destination} illustrates the real-world implementation of the database. Each beacon retains its ID, position, and rotation, ensuring multi-destination visualization while maintaining data persistence.

\section{Experiment}

To evaluate the effectiveness of our system, we conducted an experiment comparing our proposed MRNaB system with a traditional 2D system using a computer display and mouse, which is commonly used in daily life and, to our knowledge, represents a novel approach. The experiment was conducted twice, once for the 2D system and once for our MRNaB system. Fourteen participants (7 women and 7 men, aged 20 to 28) were involved, most of whom had no prior experience with VR/AR/MR or ROS. The order of the experiments for the conventional method and the proposed method was randomized.

\begin{figure}[t]
    \centering    \includegraphics[width=\columnwidth]{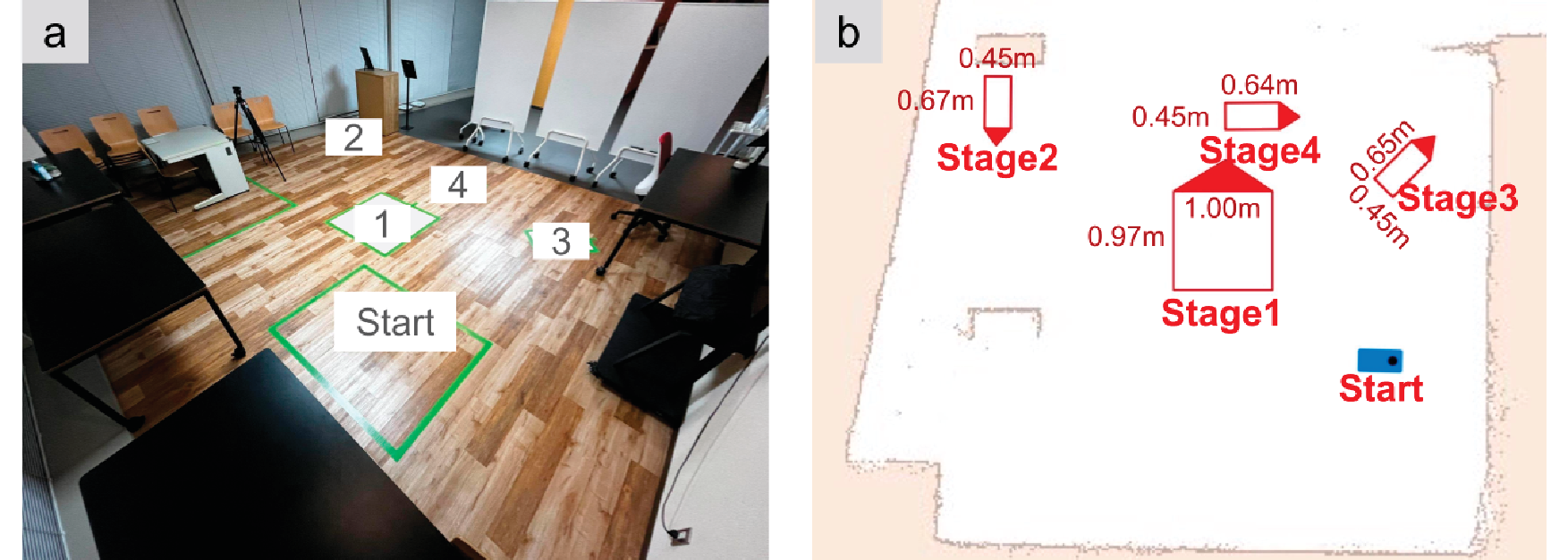}
    \caption{Environments. (a) shows the figure of the real-world experiment environment. (b) shows the figure of the experiment environment from 2D map SLAM including the area of each stage which is represented by the hollow box to represent the area and the filled triangle to represent the direction of the robot's pose in the destination. This area is not shown in the real map}
    \label{fig:Experiment Place}
\end{figure}
\begin{figure}[t]
    \centering   
    \scalebox{0.18}{
 \includegraphics{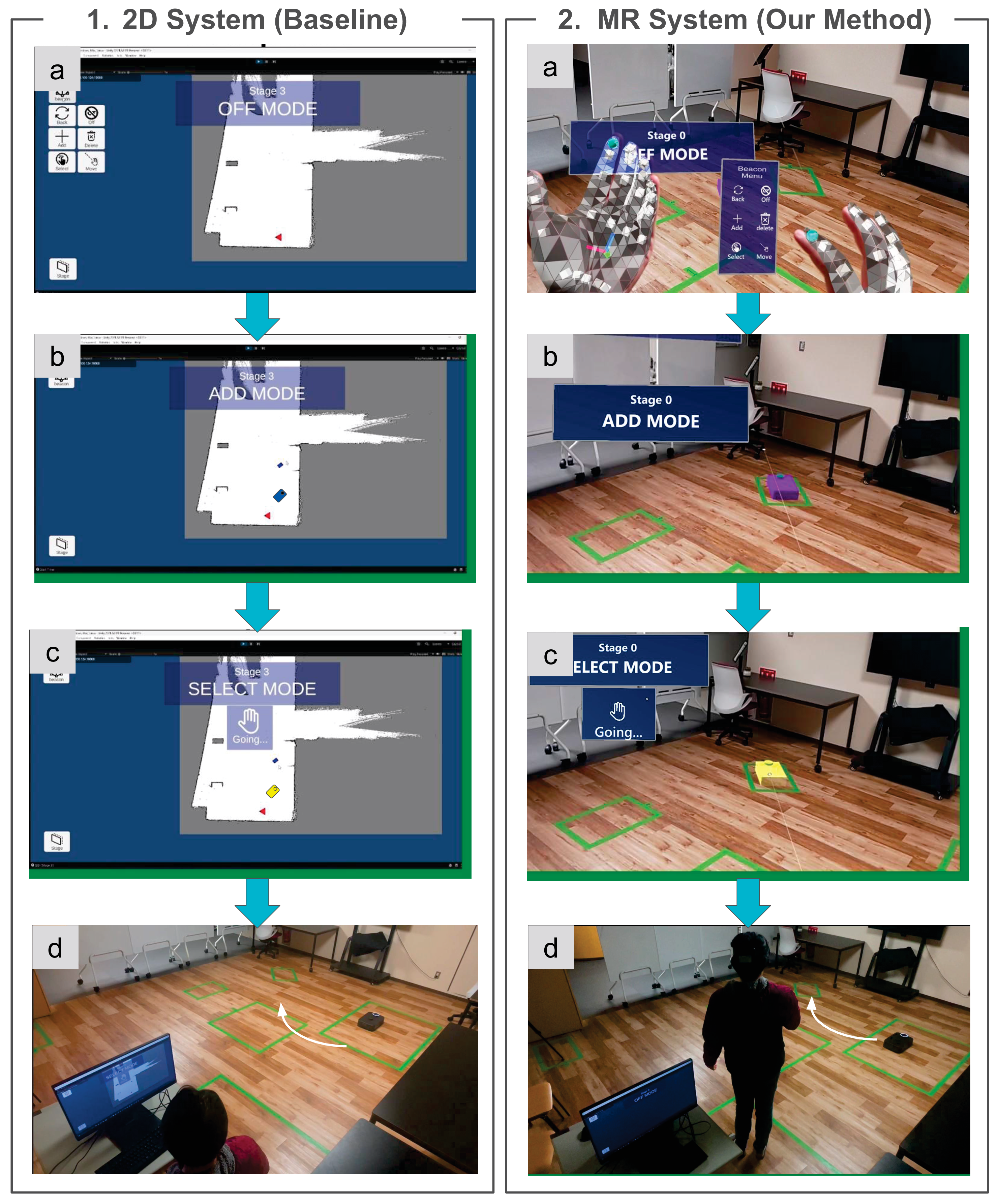}}
    \caption{Experiment Flow. The experiment was conducted twice each for 2D system as the baseline and MR system which is our proposed method. The left figure is the 2D system experiment. (1-a) Participants used beacon menu that is put on the top left of the screen to (1-b) make the beacon on the map, (1-c) do navigation by using select button, and (1-d) wait for the robot to do navigation. The right figure is our proposed method experiment. (2-a) The participant uses the beacon menu from the hand menu to (2-b) make the beacon on the real world, (2-c) do navigation by using the select button, and (2-d) wait for the robot to do navigation.}
    \label{fig:Experiment}
\end{figure} 
\subsection{Experiment Setup}

Fig.~\ref{fig:Experiment Place} shows the experiment environment. We made the environment to be similar to the household setting. For the task, we asked the participant to navigate the robot using beacon to a predetermined area with the correct direction. This area was divided into 4 stages, which were stage 1, 2, 3, and 4. 

In stage 1, we wanted to cover a reasonably spacious area, that resembled a kid's area where there is usually no clue (any reflection of object near the allowed area) from the map created by SLAM. In stage 2, we reduced the size of the area, however, we gave the opportunity for the participant to be able to see the clue which was the cabinet reflected from the map created by SLAM. This stage resembled the coffee station in the household environment. In stage 3, we kept the similar size of the area, while starting to reduce the clue reflected from the map created by SLAM, which was just a small reflection of the table . This stage resembled the working seat. Lastly in stage 4, we removed any clue from the area surrounding the target area, thus making the participant needed to use the information from the previous stage to create the MR-beacon in 2D system. This stage resembled the case where the robot needs to get the water from the water leakage from the roof.

Each of these stages was marked by green tape, and participants had to navigate the robot within the designated area. If the robot stopped or deviated outside the allowed area, which was defined as any area beyond the outer edge of the green tape, participants had to navigate the robot again using the beacon.

Fig.~\ref{fig:Experiment} shows the experiment flow.
Each of the experiment was conducted as follows: First, participants would be given an explanatory video on how to use each of the systems. After that, we helped the participants to get used to each of the main functions. Next, we let the participants to try the system themselves. Last was the real experiment. Users are required to create a beacon 
 by Add Button and then navigate robot to each stages by using Select Button. Every time the robot was navigated outside the area, participants would be required to move the beacon by using Move Button and navigate it again.

\subsection{Evaluation Indices}
Evaluation indices would be divided into 2, which were objective indices and subjective indices. For objective indices, we would first measure the total action number before navigation measurement which refers to the total number of clicks on the Add and Move buttons. This was the total action needed before participants finally be able to navigate the robot to the correct place. Second, we would measure navigation number measurement results per task which was the total navigation done by user per stage. This reflects how many times the user needed to adjust the robot’s placement before completing the task. Third, we would measure the action time measurement result. The total action time refers to the total time needed during Add and Move mode.

\begin{table}[t]
\caption{SUS Questionnaire Question List}
\label{table:SUS}
\centering
\scalebox{0.85}{
\begin{tabular}{cl}
\toprule
\multicolumn{1}{c}{No.}&\multicolumn{1}{c}{Question}\\
\midrule
1&I think that I would like to use this system frequently.\\
2&I found the system unnecessarily complex.\\
3&I thought the system was easy to use.\\
4&I think that I would need the support of a technical person to be able  \\ & to use this system. \\
5&I found the various functions in this system were well integrated. \\
6&I thought there was too much inconsistency in this system. \\
7& I would imagine that most people would learn to use this system very  \\ & quickly.  \\
8& I found the system very cumbersome to use.\\
9& I felt very confident using the system to navigate the robot.\\
10& I needed to learn a lot of things before I could get going with this system.\\
\bottomrule
\end{tabular}
}
\end{table}
For subjective indices, we measured usability using the System Usability Scale (SUS) questionnaire \cite{SUS}. The questionnaire consisted of 10 questions related to system usability, as listed in Table~\ref{table:SUS}. Odd-numbered questions were phrased in a positive context, while even-numbered questions were phrased in a negative context. Each question was rated on a scale from 0 to 4, and the total score was then multiplied by 2.5, resulting in a final score ranging from 0 to 100.

To evaluate the statistical difference, we first did Shaphiro-Wilk test \cite{shapirotest} to decide whether our data followed normal distribution or not. Since all of our data didn't follow the normal distribution, we used Wilcoxon Signed Rank Test \cite{wilcoxon} to test the $p$-value from our data.

\subsection{Experiment Result}
\subsubsection{Total Action Number Before Navigation Measurement Result}
\label{sec:totalActionNumber}

\begin{figure}[t]
    \centering
    \includegraphics[width=0.9\columnwidth]{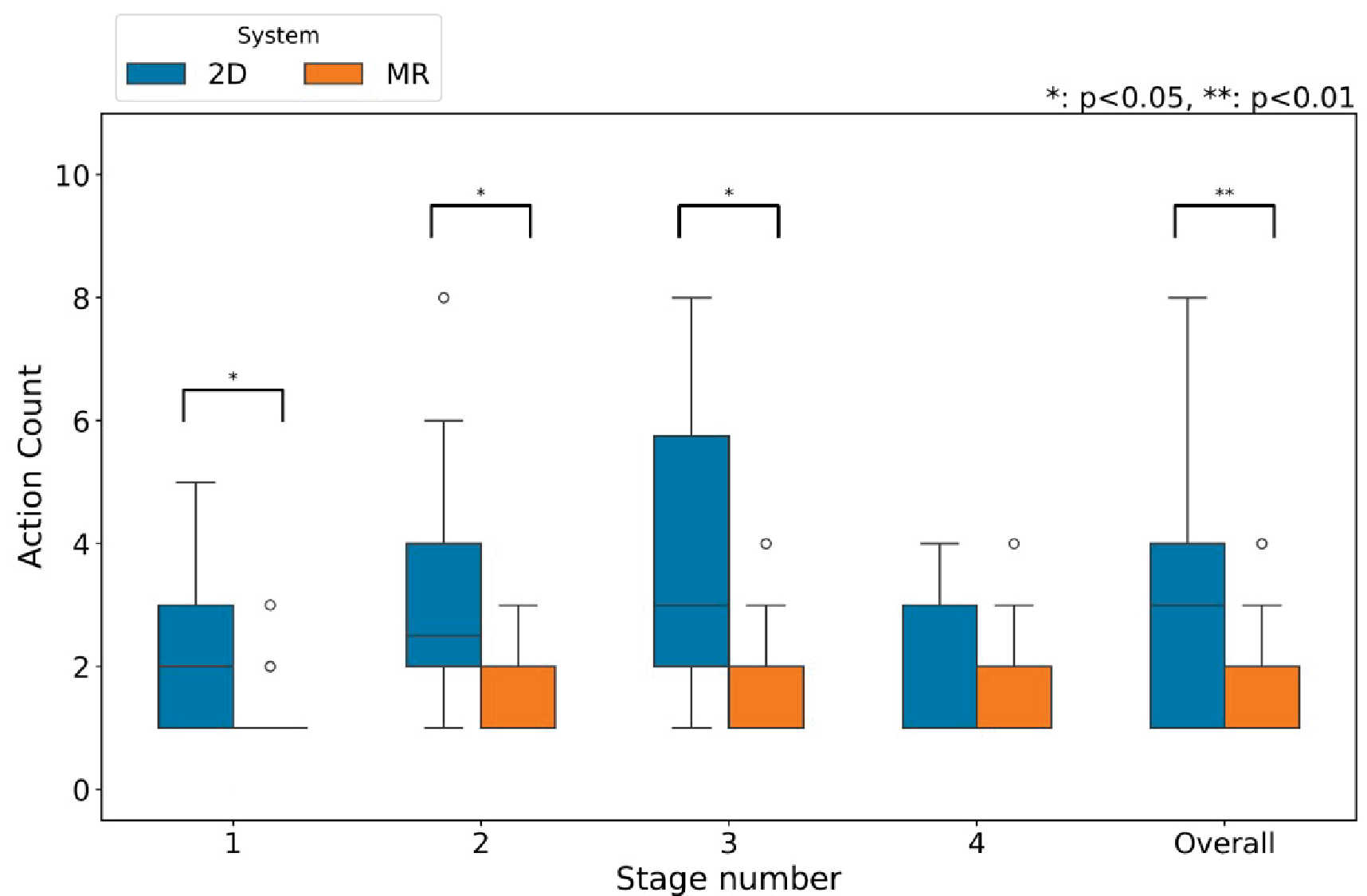}
    \caption{Total Action Number Before Navigation Measurement Result. The statistical difference can be found in stages 1, 2, 3, and overall.}
    \label{fig:total_action_before_navigation_bar}
\end{figure}

The result is shown in Fig.~\ref{fig:total_action_before_navigation_bar}. Given that the $p$-values for stages 1, 2, 3, and Overall were below 0.05, we concluded there was statistical significance across these stages. Referencing Table~\ref{tab:effectiveness_comparison}, it was evident that participants required less beacon actions to navigate the robot accurately using our system compared to a traditional 2D system where our system only required 1.59 tries on average while the 2D system required 2.95 tries on average. We posit that these changes came because, in our system, participants were able to navigate the robot directly in the real world compared to the 2D system thus making it easier for participants to navigate the robot directly to the desired location.

\begin{table}[t]
\centering
\caption{Effectiveness Comparison by Action Number}
\scalebox{1}{
\begin{tabular}{ccc}
\toprule
Stage Number & Baseline (2D) & MRNaB  \\ 
\midrule
1            & 2.43 & \textbf{1.29}   \\ 
2            & 3.14 & \textbf{1.43}  \\ 
3            & 3.79 & \textbf{1.93}  \\ 
4            & 2.43 & \textbf{1.71} \\ 
\midrule
Overall      & 2.95 & \textbf{1.59}  \\ 
\bottomrule
\end{tabular}
}
\label{tab:effectiveness_comparison}
\end{table}

\begin{figure}[t]
    \centering
    \includegraphics[width=0.9 \columnwidth]{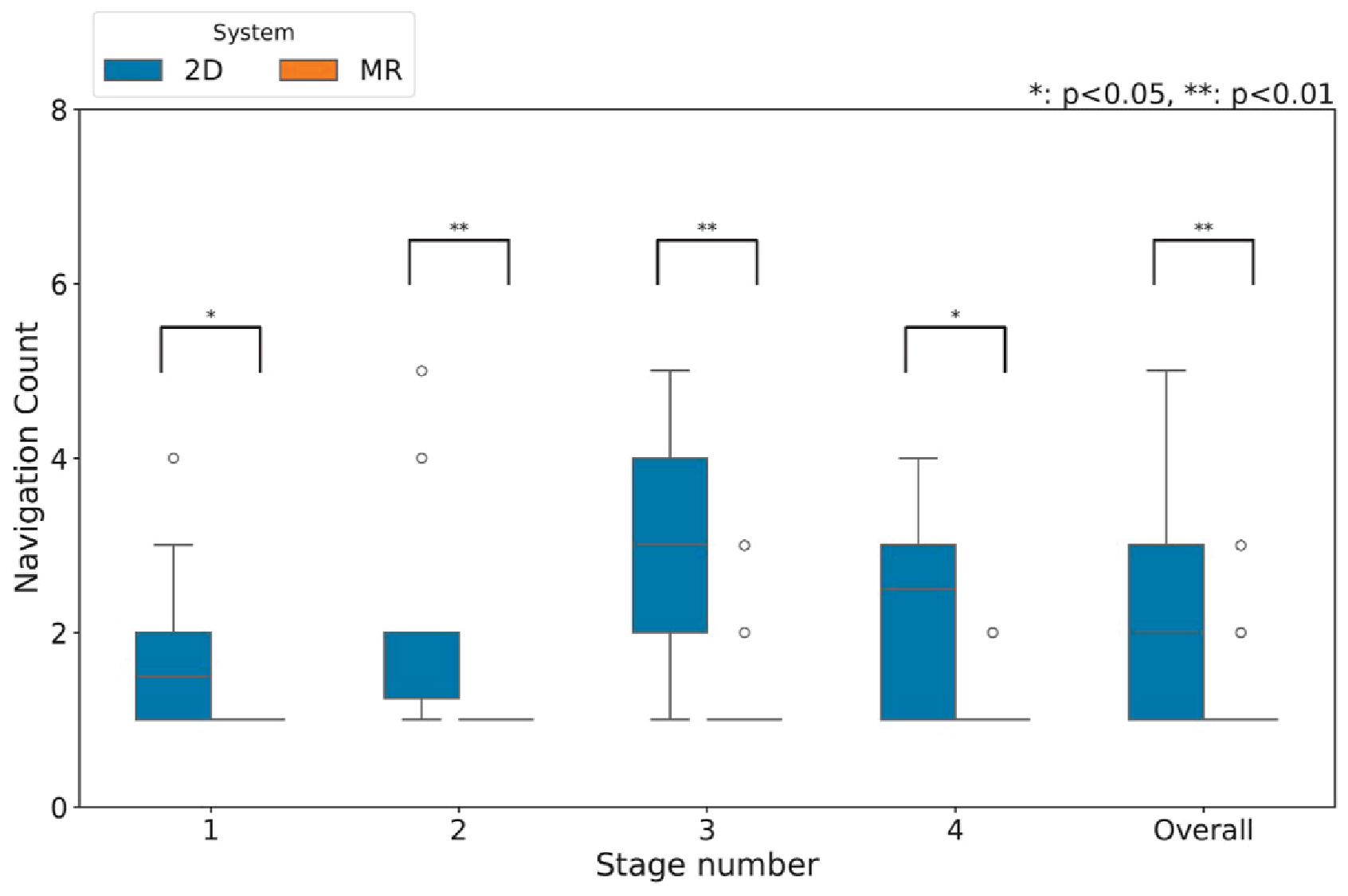}
    \caption{Navigation Number Measurement Result. Statistical differences can be found in stages 1, 2, 3, 4,  and overall.}
    \label{fig:navigation_bar}
\end{figure}

\begin{table}[t]
\centering
\caption{Effectiveness Comparison by Navigation Number}
\scalebox{1}{
\begin{tabular}{ccc}
\toprule
Stage Number & Baseline (2D) & MRNaB  \\ 
\midrule
1            & 1.79 & \textbf{1.00}  \\ 
2            & 2.07 & \textbf{1.00}  \\ 
3            & 3.14 & \textbf{1.36}  \\ 
4            & 2.21 & \textbf{1.21}  \\ 
\midrule
Overall      & 2.30 & \textbf{1.14}  \\ 
\bottomrule
\end{tabular}
}
\label{tab:effectiveness_comparison2}
\end{table}

\begin{figure}[t]
    \centering
    \includegraphics[width=0.9 \columnwidth]{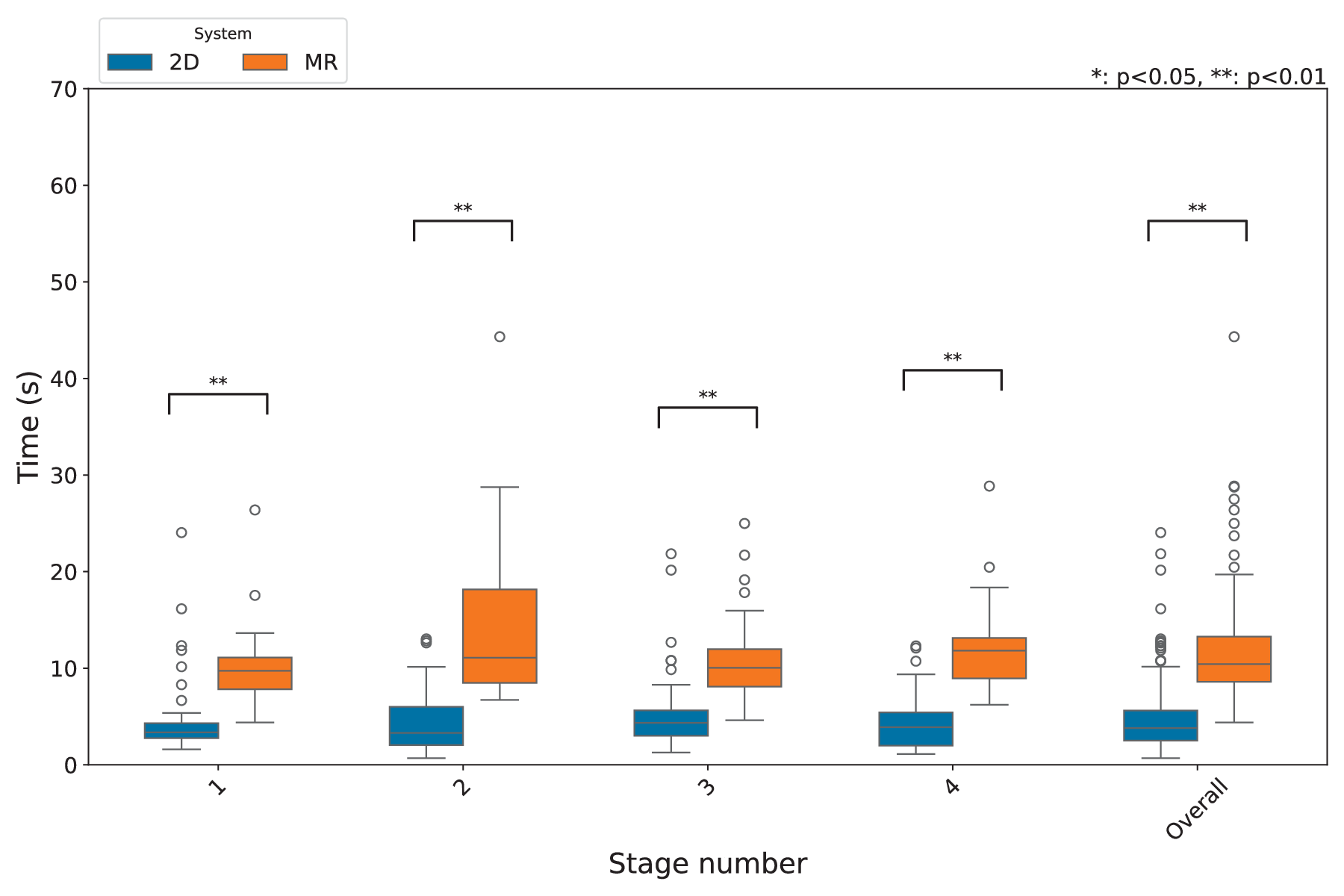}
    \caption{Action Time Measurement Result. The statistical difference can be found in stages 1, 2, 3, 4, and overall.}
    \label{fig:Time_box}
\end{figure}

\begin{table}[t]
\centering
\caption{Effectiveness Comparison by Action Time}
\scalebox{1}{
\begin{tabular}{ccc}
\toprule
Stage Number & Baseline (2D) & MRNaB  \\ 
\midrule
1            & \textbf{9.13} & 14.36  \\ 
2            & \textbf{9.84} & 18.79  \\ 
3            & \textbf{8.37} & 13.38  \\ 
4            & \textbf{8.87} & 14.46  \\ 
\midrule
Overall      & \textbf{9.05} & 15.25  \\ 
\bottomrule
\end{tabular}
}
\label{tab:effectiveness_comparison3}
\end{table}

\begin{figure}[t]
    \centering
    \includegraphics[width=0.75 \columnwidth]{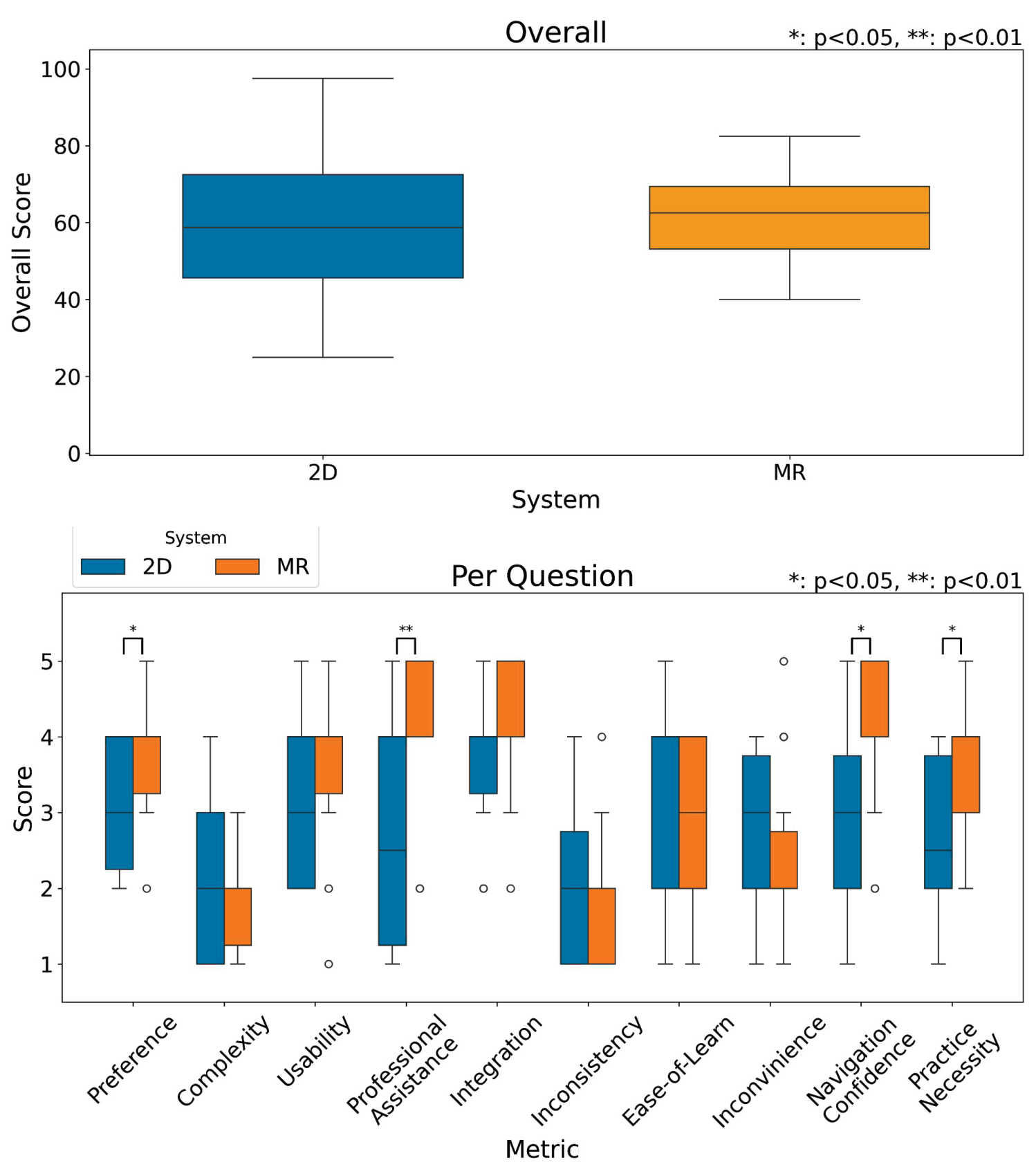}
    \caption{SUS Result. The top figure shows the overall score of the SUS for 2D and MR systems where statistical differences can't be found. The bottom figure shows the score for each of the questions for 2D and MR systems.}
    \label{fig:SUS_box_overall}
\end{figure}

\subsubsection{Navigation Number Measurement Results Per Task}
\label{sec:Navigation Number Result}
The result is shown in Fig.~\ref{fig:navigation_bar}.
Given that the $p$-values for all stages were lower than 0.05, we concluded that there is statistical significance in the data presented. Referencing Table~\ref{tab:effectiveness_comparison2}, it was evident that participants required less navigation for the robot to reach its destination using our system compared to the traditional 2D system, where our system only required 1.14 tries on average while the 2D system required 2.30 tries on average.  We posit that these differences came because in our system, participants were able to navigate the robot directly in the real world compared to the 2D system where participants needed to shift their attention and did the calculation based on the clue from the map.

\subsubsection{Action Time Measurement Result}

The result is shown in Fig.~\ref{fig:Time_box}.
Given that the $p$-values for all stages were lower than 0.05, we concluded that there is statistical significance in the data presented. Referencing Table~\ref{tab:effectiveness_comparison3}, it was evident that participants required less action time to complete the task using traditional 2D system compared to our system. Our system only required 15.25 seconds action time on average, while the 2D system required 9.05 seconds action time on average. We posit that these differences came because in our system participants could see the targeted location in the real world, leading them to take more time to place it accurately. In contrast, the traditional system did not provide a visible target location, causing participants to place the beacon more quickly.

However, despite these drawbacks, as observed in Section~\ref{sec:totalActionNumber} and Section~\ref{sec:Navigation Number Result}, users performed fewer beacon-related actions and navigation attempts, suggesting improved efficiency in directing the robot to its destination. This indicates that while our system required more time per action, it ultimately reduced redundant interactions, enabling users to achieve their goals with fewer overall steps.

\subsubsection{Subjective Evaluation of Usability}
The result for the overall score and the score per question are shown in Fig.~\ref{fig:SUS_box_overall}.
For the individual questions, statistical significance was found only in preference, professional assistance, navigation confidence, and practice necessity, suggesting that our system was generally user-friendly and facilitated easy navigation of the robot. We posit that the ability to see the beacon throughout the entire process increased the confidence and preference of users. Additionally, we posit that viewing detailed environmental information, rather than the abstract representation provided by a 2D map, led to higher scores for these indices.
 
 However, as most participants lacked experience with MR head-mounted displays (HMDs), particularly the ``air tap'' gesture, a learning curve was evident. This unfamiliarity with the device necessitated a period of adaptation before users could comfortably utilize the system, thus resulting the higher score for professional assistance and practice necessity. For the remaining questions, the variability in how quickly individuals learned the ``air tap'' gesture introduced a subjective element to the responses, potentially contributing to the lack of statistical significance observed in these cases which contributed further to no statistical significance for the overall score.

Thus, the effectiveness of our method can be seen through comprehensive real-world experiments.

\section{Conclusion}
This paper has proposed MRNaB, an interface for navigating robot by using MR-beacon with Hololens~2. The establishment is done by using ``air tap'' gesture with 4 functionalities which are ``Add'', ``Move'', ``Select'', and ``Delete''. MRNaB is also integrated with a database system, eliminating the need for users to create a new MR-beacon each time the project is restarted. This ensures multi-destination visualization and maintains data persistence across sessions. Our experiments comparing MRNaB with a traditional 2D system demonstrated that despite users taking longer time to place the beacon accurately, MRNaB required less beacon settings frequency which significantly decreased the amount of navigation needed. MRNaB was also proven to increase user's confidence in order to navigate the robot to the right direction. 

There are still a few limitations to the current work. For instance, occlusion can lead to optical illusions that may mislead users when utilizing our proposed method. Additionally, establishing the MR-beacon at very far distance is challenging because the decreasing angle between the viewpoint and the floor makes it difficult to specify the location. We plan to address these issues in future work by implementing solutions for occlusion and incorporating a minimap for far-distance establishment.


\begin{thebibliography}{10}
\providecommand{\url}[1]{\normalfont{#1}}
\providecommand{\urlprefix}{Available from: }

\bibitem{journal:ros}
Quigley~M, Conley~K, Gerkey~B, et~al. {ROS: an open-source Robot Operating System}. Vol.~3; 2009.

\bibitem{ROSUnder}
Ngo~AT, Tran~NH, Ton~TP, et~al. Simulation of hybrid autonomous underwater vehicle based on ros and gazebo. In: 2021 International Conference on Advanced Technologies for Communications (ATC); 2021. p. 109--113.

\bibitem{ROSWheelchair}
Cruz~AB, Sousa~A, Reis~LP. Controller for real and simulated wheelchair with a multimodal interface using gazebo and ros. In: 2020 IEEE International Conference on Autonomous Robot Systems and Competitions (ICARSC); 2020. p. 164--169.

\bibitem{ROSNeuro}
Beraldo~G, Castaman~N, Bortoletto~R, et~al. Ros-health: An open-source framework for neurorobotics. In: 2018 IEEE International Conference on Simulation, Modeling, and Programming for Autonomous Robots (SIMPAR); 2018. p. 174--179.

\bibitem{ROShip}
Velamala~SS, Patil~D, Ming~X. Development of ros-based gui for control of an autonomous surface vehicle. In: 2017 IEEE International Conference on Robotics and Biomimetics (ROBIO); 2017. p. 628--633.

\bibitem{journal:arviz}
Hoang~KC, Chan~WP, Lay~S, et~al. {ARviz: An Augmented Reality-Enabled Visualization Platform for ROS Applications}. IEEE Robotics \& Automation Magazine. 2022;\hspace{0pt}29(1):58--67.

\bibitem{Salvato2022mr}
Salvato~M, Heravi~N, Okamura~AM, et~al. Predicting hand-object interaction for improved haptic feedback in mixed reality. IEEE Robotics and Automation Letters. 2022;\hspace{0pt}7(2):3851--3857.

\bibitem{Devo2022mr}
Devo~A, Mao~J, Costante~G, et~al. Autonomous single-image drone exploration with deep reinforcement learning and mixed reality. IEEE Robotics and Automation Letters. 2022;\hspace{0pt}7(2):5031--5038.

\bibitem{AR4Robot}
Makhataeva~Z, Varol~HA. Augmented reality for robotics: A review. Robotics. 2020;\hspace{0pt}9(2).

\bibitem{Fan2023mr}
Fan~W, Guo~X, Feng~E, et~al. Digital twin-driven mixed reality framework for immersive teleoperation with haptic rendering. IEEE Robotics and Automation Letters. 2023;\hspace{0pt}8(12):8494--8501.

\bibitem{Penco2024mr}
Penco~L, Momose~K, McCrory~S, et~al. Mixed reality teleoperation assistance for direct control of humanoids. IEEE Robotics and Automation Letters. 2024;\hspace{0pt}9(2):1937--1944.

\bibitem{Zhang2023mr}
Zhang~C, Lin~C, Leng~Y, et~al. An effective head-based hri for 6d robotic grasping using mixed reality. IEEE Robotics and Automation Letters. 2023;\hspace{0pt}8(5):2796--2803.

\bibitem{eyeoperation}
Lee~J, Lim~T, Kim~W. {Investigating the Usability of Collaborative Robot Control Through Hands-Free Operation Using Eye Gaze and Augmented Reality}. In: Proc. of the IEEE/RSJ International Conference on Intelligent Robots and Systems; 2023. p. 4101--4106.

\bibitem{journal:drag}
Chen~J, Sun~B, Pollefeys~M, et~al. {A 3D Mixed Reality Interface for Human-Robot Teaming}. arXiv preprint arXiv:231002392. 2023;\hspace{0pt}Available: arXiv:2310.02392 [cs.RO].

\bibitem{journal:gaze}
Zhang~G, Zhang~D, Duan~L, et~al. {Accessible Robot Control in Mixed Reality}. arXiv preprint arXiv:230602393. 2023;\hspace{0pt}Available: arXiv:2306.02393 [cs.RO].

\bibitem{gazebo}
Koenig~N, Howard~A. {Design and use paradigms for Gazebo, an open-source multi-robot simulator}. In: Proc. of the IEEE/RSJ International Conference on Intelligent Robots and Systems; Vol.~3; 2004. p. 2149--2154 vol.3.

\bibitem{MRDS}
Cepeda~JS, Chaimowicz~L, Soto~R. {Exploring Microsoft Robotics Studio as a Mechanism for Service-Oriented Robotics}. In: Proc. of the Latin American Robotics Symposium and Intelligent Robotics Meeting; 2010. p. 7--12.

\bibitem{webots}
Michel~O. {WebotsTM: Professional Mobile Robot Simulation}. International Journal of Advanced Robotic Systems. 2004;\hspace{0pt}1.

\bibitem{Web-based}
Rajapaksha~DD, Mohamed~Nuhuman~MN, Gunawardhana~SD, et~al. {Web Based User-Friendly Graphical Interface to Control Robots with ROS Environment}. In: Proc. of the International Conference on Information Technology Research; 2021. p. 1--6.

\bibitem{web-based3}
Fernando~WAM, Jayawardena~C, Rajapaksha~UUS. {Developing A User-Friendly Interface from Robotic Applications Development}. In: Proc. of the International Research Conference on Smart Computing and Systems Engineering; Vol.~5; 2022. p. 196--204.

\bibitem{web-based4}
Aarizou~M. {ROS-based web application for an optimized multi-robots multi-users manipulation}. In: Proc. of the National Conference in Computer Science Research and its Applications; 2023.

\bibitem{rviz}
Kam~H, Lee~SH, Park~T, et~al. {RViz: a toolkit for real domain data visualization}. Telecommunication Systems. 2015;\hspace{0pt}60:1--9.

\bibitem{web-based2}
Tiddi~I, Bastianelli~E, Bardaro~G, et~al. {A User-friendly Interface to Control ROS Robotic Platforms}. In: Proc. of the International Workshop on the Semantic Web; 2018.

\bibitem{surrogate}
Walker~ME, Hedayati~H, Szafir~D. Robot teleoperation with augmented reality virtual surrogates. In: 2019 14th ACM/IEEE International Conference on Human-Robot Interaction (HRI); 2019. p. 202--210.

\bibitem{map2}
Wu~M, Dai~SL, Yang~C. {Mixed Reality Enhanced User Interactive Path Planning for Omnidirectional Mobile Robot}. Applied Sciences. 2020;\hspace{0pt}10(3):1135.

\bibitem{thruWall}
Gu~M, Cosgun~A, Chan~WP, et~al. {Seeing Thru Walls: Visualizing Mobile Robots in Augmented Reality}. In: Proc. of the IEEE International Conference on Robot \& Human Interactive Communication. IEEE; 2021.

\bibitem{intent}
Walker~M, Hedayati~H, Lee~J, et~al. {Communicating Robot Motion Intent with Augmented Reality}. In: Proc. of the ACM/IEEE International Conference on Human-Robot Interaction; 2018. p. 316--324.

\bibitem{ihandai}
Owaki~K, Techasarntikul~N, Shimonishi~H. {Human Behavior Analysis in Human-Robot Cooperation with AR Glasses}. In: Proc. of the IEEE International Symposium on Mixed and Augmented Reality; Los Alamitos, CA, USA. IEEE Computer Society; 2023. p. 20--28.

\bibitem{wheelchair}
Zolotas~M, Demiris~Y. {Towards Explainable Shared Control using Augmented Reality}. In: Proc. of the IEEE/RSJ International Conference on Intelligent Robots and Systems; 2019. p. 3020--3026.

\bibitem{wheelchair2}
Zolotas~M, Elsdon~J, Demiris~Y. {Head-Mounted Augmented Reality for Explainable Robotic Wheelchair Assistance}. In: Proc. of the IEEE/RSJ International Conference on Intelligent Robots and Systems; 2018. p. 1823--1829.

\bibitem{multiuser}
Regal~F, Petlowany~C, Pehlivanturk~C, et~al. {AugRE: Augmented Robot Environment to Facilitate Human-Robot Teaming and Communication}. In: Proc. of the IEEE International Conference on Robot and Human Interactive Communication; 2022. p. 800--805.

\bibitem{joystickoperation}
Kot~T, Novák~P, Bajak~J. {Using HoloLens to create a virtual operator station for mobile robots}. In: Proc. of the International Carpathian Control Conference; 2018. p. 422--427.

\bibitem{Drone}
Angelopoulos~A, Hale~A, Shaik~H, et~al. {Drone Brush: Mixed Reality Drone Path Planning}. In: Proc. of the ACM/IEEE International Conference on Human-Robot Interaction; 2022. p. 678--682.

\bibitem{holo-spok}
Quesada~RC, Demiris~Y. {Holo-SpoK: Affordance-Aware Augmented Reality Control of Legged Manipulators}. In: Proc. of the IEEE/RSJ International Conference on Intelligent Robots and Systems; 2022. p. 856--862.

\bibitem{sar}
Cruz~C, Cerro~J, Barrientos~A. {Mixed-reality for quadruped-robotic guidance in SAR tasks}. Journal of Computational Design and Engineeringy. 2023;\hspace{0pt}10.

\bibitem{SUS}
Brooke~J. {SUS: A quick and dirty usability scale}. Usability Eval Ind. 1995;\hspace{0pt}189.

\bibitem{shapirotest}
Shapiro~SS, Wilk~MB. {An Analysis of Variance Test for Normality (Complete Samples)}. Biometrika. 1965;\hspace{0pt}52(3/4):591--611.

\bibitem{wilcoxon}
Rey~D, Neuhäuser~M. {Wilcoxon-Signed-Rank Test}. International Encyclopedia of Statistical Science. 2011;\hspace{0pt}:1658--1659.

\end{thebibliography}
\end{document}